\documentclass[a4paper, 10pt, conference]{ieeeconf}
\IEEEoverridecommandlockouts                              

\usepackage{cite}
\usepackage{amsmath,amssymb,amsfonts}
\usepackage{algorithmic}
\usepackage{graphicx}
\usepackage{textcomp}
\usepackage{xcolor}

\usepackage[ruled,linesnumbered ]{algorithm2e}
\usepackage{subfig}                         
\usepackage{multirow, multicol}
\usepackage{float}

\title{\LARGE \bf
What is the appropriate speed for an autonomous vehicle? \\ Designing a Pedestrian Aware Contextual Speed Controller}

\author{Daniel Jiang, Stewart Worrall and Mao Shan
\thanks{The authors are with the Australian Centre for Field Robotics, The University of Sydney, Sydney, NSW 2006, Australia.
        E-mails: {\tt\footnotesize  djia3099@uni.sydney.edu.au, \{stewart.worrall, mao.shan\}@sydney.edu.au}}%
}

\begin{document}

\maketitle

\begin{abstract}
Social acceptance is a major hurdle for autonomous vehicle technology, central to which is ensuring both passengers and nearby pedestrians feel safe.
This idea of `feeling safe' and perceived safety is highly subjective and rooted in human intuition. As such, traditional analytical approaches to autonomous navigation often fail to cater for the social expectations of individuals. Therefore, this paper proposes an approach to capture the complexity of social expectations and integrate this complexity into a 3-layered Contextual Speed Controller. The layers were; the legal road speed limit, the socially acceptable speed given the number of nearby pedestrians, and the socially acceptable speed based on proximity to nearby pedestrians. 
An implementation of this layered approach was tested in areas of both low and high vehicle-pedestrian interactions. From the experiments conducted, the lower two layers were seen working in tandem to modulate the vehicle speed to appropriate levels that mimicked conservative human driver behaviour.
In summary, this work quantified the relationship between pedestrian context and socially acceptable vehicle speeds, allowing for more perceivably safe autonomous driving. Furthermore, the need for different driving schemes for navigating different road environments was identified. 

\end{abstract}

\section{Introduction}
\begin{figure}[h]
    \centering
    \subfloat[][Regular Roads\label{}]
        {\includegraphics[height = 2.8cm]{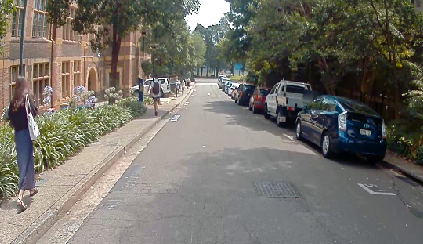}}
    \subfloat[][Shared Spaces\label{}]
        {\includegraphics[height = 2.8cm]{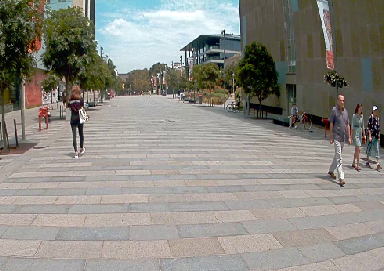}}
    \caption{Example Contexts}
    \label{examples}
\end{figure}

Ensuring social acceptance is a major challenge for the widespread deployment of autonomous vehicles. As it stands, human drivers intuitively interpret social cues and contexts to navigate roads and close pedestrian interactions. An easy example demonstrating this is the appropriateness of driving at 60 KPH. On a major road, nearby pedestrians on the sidewalk will likely be unfazed by a vehicle passing closeby at this high speed. If this were however in a shared space such as a busy plaza, a high speed vehicle may be perceived as dangerous and cause significant discomfort. An example of these contexts can be seen in Figure \ref{examples}. This perception of safety is difficult to emulate and largely unexplored, leading to more instances of the latter example. Thus, the focus of this paper is to discuss ways in which human intuition can be incorporated into vehicle speeds and increase the social acceptance of this developing technology. 

Currently, path planners are capable of processing dynamic situations in realtime. This is often done by either allowing a buffer zone for dynamic objects such as pedestrians, or by making some prediction of their actions and planning accordingly. The issue here is that these systems predominantly only plan a path, and the vehicle speed is a consequence of the path planned, taking no heed of the surrounding pedestrian expectations. Incorporating both speed and path planning into one system increases the computational complexity and reduces the frequency at which the planner can operate, due to the higher dimensionality of the problem. 

Therefore, this paper proposes the concept of a `Contextual Speed Controller' (so forth referred to simply as Speed Controller), which acts as a dedicated speed modulator for responding to surrounding pedestrians. By decoupling the behaviour of the vehicle around pedestrians from the rest of the system, overall runtimes can be substantially reduced whilst also creating a safer and more comfortable environment as perceived by surrounding pedestrians. It is envisaged that this Speed Controller is strictly modular and can be implemented in any autonomous driving system to improve comfort in low-speed contexts where pedestrians may be present. As a proof of concept, a Speed Controller was successfully implemented for vehicles from the University of Sydney's (USyd's) Intelligent Transport Systems (ITS) group and tested on recorded data.

The proposed Speed Controller operates in three distinct layers to limit the speed of the vehicle whilst considering the context of the road: 
\begin{enumerate}
    \item (Legal) What is the legal speed limit of the current road? 
    \item (Context) What is an appropriate speed based on the number of pedestrians in the current traffic context?
    \item (Proximity) What is an appropriate speed based on the proximity of pedestrians to the vehicle path?
\end{enumerate}
Identifying the legal speed limit of the road is a widely investigated topic, thus this paper instead focuses on discussing the latter two questions. 

The question: ``What is an appropriate speed given the number of nearby pedestrians in the current traffic context?'' is widely un-investigated, with most discussion relating to the correlation between collision speed and risk of injury. This however, fails to address the human complexity of the problem and neglects perceptions of safety. 
In this paper, it is proposed that by performing a quantitative analysis of a human operating the vehicle in different contexts and spaces, a model can be developed to map contexts to appropriate vehicle speeds. This model could then be incorporated into the navigation stack of an autonomous vehicle and inform future decisions. 

It is also necessary to consider how the proximity of pedestrians to the vehicle path affects the vehicle speed. One approach to tackle this is to consider the location of people in 3D and calculate their position as a function of distance \textit{from} the path and distance  \textit{along} the path. Pedestrians near the vehicle path are projected onto the path at a further distance based on a `lateral scaling factor' that can be tuned to adjust how conservative the system will behave. This accounts for the uncertainty of dynamic objects and allows pedestrians both on and off the path to be evaluated in the same way. An appropriate speed is estimated based on allowable braking distance. This paper also presents a method of achieving realtime 3D pedestrian detection through a fusion of 3D pointcloud clusters and 2D pedestrian bounding boxes (bboxes). 

Upon combining these three layers, experiments conducted showed that this Speed Controller concept is capable of realtime performance and qualitative analysis showed vehicle behaviour that mimicked conservative human driving. The experiments were conducted on a variety of data recording playbacks, comparing human control to the implemented system.

The contributions of this paper can be summarised as: 
\begin{itemize}
    \item A proposal for a layered approach to modulating vehicle speed for improved pedestrian comfort. The feasibility and benefits this approach are analysed. 
    \item A demonstration of the use of a scaling factor to adjust the conservative-ness of such a Speed Controller system.
    \item An identification of the need for different driving schemes to be applied in different contexts, and quantifying one such approach.
\end{itemize}

\section{Related Work}\label{sec:lit_review}
Very few investigations have been conducted into assessing the relationship between autonomous vehicles and pedestrians with regard to appropriate speeds. While some papers address other branches of this problem, such as whether people regard self-driving vehicles as safe holistically \cite{perception2}, or how different factors on the road may affect pedestrian behaviour \cite{perception1}, none were found to quantify the speeds that pedestrians found comfortable in vehicle-pedestrian interactions. This is not to say however, that this topic isn't worth investigating as reports such as one by the U.S. Department of Transport \cite{usTransport} discussed in detail how people felt about vehicles driving at certain speeds and in certain volumes along a highway. Highlighting the highly subjective and varying nature of perceptions of safety, this report found the comfort of individuals dropped by 62\% when the platoon increased from 5 to 25 vehicles. Similarly, a paper by Pelikan \cite{pelikan} highlights the difficulty of communicating intention to other users of the road as an autonomous vehicle. A particular case study by Pelikan found the investigated autonomous bus needing to abruptly break more often than normal due to an inability to process social cues.

Much of the existing quantitative research into safe speeds for vehicle-pedestrian interactions is with respect to injury level and risk of death \cite{kroyer,Rosen2009,usTransport}. Again, this still does not address human perceptions of safety and other social cues that may influence vehicle behaviour. Therefore, further investigations are required to determine these relationships.

This investigation into close vehicle-pedestrian interactions is further motivated by the growing interest in shared spaces globally. Following successful demonstrations of the shared space, or `woonerf', concept by Mondermann \cite{Woonerfgoed,NicoleThomas2014}, more city planners have begun utilising these shared spaces to promote community engagement, user accessibility and overall safety\cite{nsw_ss, winnipeg_ss}. One recent example is in New South Wales, Australia, where the government sought to introduce more shared spaces to improve public areas and facilitate exercise \cite{nsw_ss}. Research into appropriate speeds for these shared spaces has also been limited, with only one fairly dated example by Gilles \cite{UNSWShared}. Thus, further investigation is required if the development of self-driving vehicles is to evolve alongside modern cities.

Pedestrian detection is another challenging aspect of autonomous vehicle development that must be performed in order to evaluate the pedestrian context of the vehicle's surroundings. While detectors in 2D have demonstrated very promising results whilst running in realtime \cite{kitti}, accurate detection in 3D is still lagging behind. This is problematic as 3D detection is required to determine the distance to a target, such that time to collision can be computed accurately and risk of collision minimised. 

Cutting-edge detectors in 3D include those by Liu et al. \cite{pcsTemplate} and Spinello et al. \cite{layered}, both of which had high precision and recall rates. However, despite using different pointcloud processing techniques, both achieve a detection frequency of around 1 Hz, far from realtime. Looking at top performers for the KITTI dataset \cite{kitti}, detection rates are inadequate, lying around 50\%, though demonstrated much faster detection rate of 25 Hz, sufficient for realtime operation. This balancing act in 3D detection performance suggests room for further development of existing techniques. 

An alternative approach to detecting pedestrians in 3D is to use a so-called `fusion' approach, where multiple modes of data (such as 2D RGB images and 3D pointclouds) are combined into one. The resultant data is richer in information and can lead to higher accuracies, with machine learning approaches commonly employed to process such data. Often however, the higher dimensionality or the additional steps of fusion cause longer computational times, as is shown in recent literature \cite{fusionModes, tosteberg, mvRF,pointFusion,crossFusion}. Thus, in order to achieve realtime performance, alternative and perhaps more efficient approaches to fusion should be taken. 

\begin{figure*}[ht]
\begin{center}
    \includegraphics[width=0.8\linewidth]{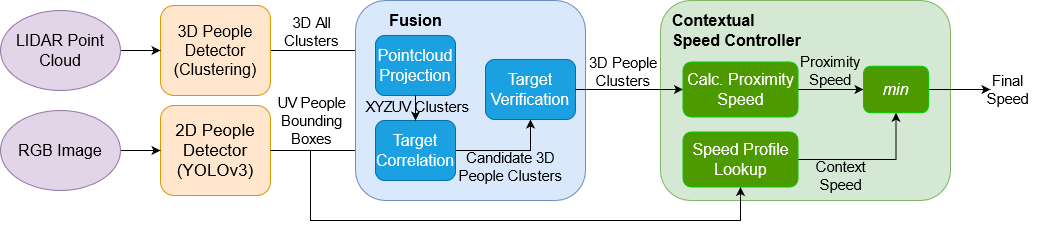}
\end{center}
       \caption{Contextual Speed Controller Pipeline}
    \label{pipeline}
\end{figure*}%

\section{Dataset}
The methods and experiments detailed in this paper use the naturalistic urban driving data contained in the USyd Campus Dataset \cite{usydDataset}. This dataset consists of ROS Bags recorded by a human-driven electric vehicle driving through a fixed route weekly for over one year. The USyd Campus Dataset was chosen for its comprehensive suite of sensors, diverse range of operational scenarios and environments, along with readily available tools. Importantly, the dataset contains odometry, RGB-camera and LIDAR data, in addition to driving through roads and shared pedestrian areas of various pedestrian density. The varying conditions allow for more robust testing and evaluation of the system proposed in this paper through playback of the recorded data.

\section{Context Layer}
Here, an approach to determining an appropriate vehicle speed given the pedestrian density within the visible scene is presented. This is achieved by creating a model, or `Speed Profile', that maps pedestrian densities to suitable speeds. To best resolve the complexity of close human-vehicle interactions, a human driver is tasked with driving through various pedestrian densities. Their behaviour is then taken as reference for `suitable speeds' and used to inform the Contextual Speed Controller.

\subsection{Obtaining Data} \label{ssec:yolo}
Two key pieces of information required for the Speed Profile, the vehicle speeds and the number of pedestrians in the scene. The vehicle speed was directly obtained from the recorded onboard odometry. Detecting pedestrians however, required the use of a 2D object detector. YOLOv3 \cite{yolov3} was chosen for this paper, specifically, a version of Darknet \cite{darknet} that was wrapped into a ROS node. While other detectors have been shown to perform better/as fast as YOLO for certain datasets, this detector was chosen due to the simplicity of implementation. With an out-of-the-box pre-trained model, high detection rates are achieved whilst maintaining low inference times ($<60$ ms on NVIDIA GTX1060), adequate for the purposes of this research. 

As a generic object detector, only bounding boxes with class `person' were considered. However, significant detail of each bbox was recorded, including; total number of detections, size of each bbox and where each detection was. This allowed for considerations of location in the scene and distance from the vehicle, both factors that can affect human perceptions of safety. 

One other consideration is the mitigation of other variables that may influence driver behaviour. Examples of this include turns, static obstacles and speed bumps, all of which would cause the driver to slow down independently from the pedestrian density. Therefore, each dataset sample is also broken down into several `road types', and only informative sub-samples utilised. The useful road types categorised were: 
\begin{itemize}
    \item Shared spaces: Roads with no distinct segregation of people and vehicles. Some turns and no speed bumps. Very high pedestrian density. 
    \item Semi-shared roads: Roads with heavy vehicle-pedestrian interaction, but with distinct indication of where vehicles should be driving. No major turns or speed bumps. Moderate pedestrian density. 
    \item Regular roads: Vehicle dominant roads, usually with pedestrians only on the sidewalks. Turns and speed bumps minimised. Low pedestrian density.
\end{itemize}

\subsection{Processing Data}
Firstly, all the analog speed data was discretised into smaller intervals of 5 KPH. This discretised data was then tallied to determine the number of instances of each speed for each pedestrian density. The resultant binned data could then be displayed as a stacked bar graph for interpretation. 

A lookup table was produced by taking the mean speed of each pedestrian density bin. In practice, as the vehicle navigates a space and detects pedestrians in the scene, this lookup table can be referred to for an appropriate conservative maximum speed for the vehicle to operate at. 

\section{Proximity Layer}
The proximity of pedestrians to the intended vehicle path is another important consideration when modulating vehicle speed. To consider this, pedestrian detection is first required. As outlined in Section \ref{sec:lit_review}, machine learning approaches typically increase the latency of the vehicle's response to sudden movements from pedestrians. Instead, a fusion stage is implemented where 3D pointclouds are projected onto an image and matched against bounding boxes containing people for classification. This approach was chosen for its fast execution, ease of implementation and acceptable performance.

In this section, the detection technologies used as inputs to the fusion stage are introduced, followed by the fusion stage itself. After, the details of how the 3D detections are used to compute an appropriate speed are outlined.

The pipeline is illustrated in Figure \ref{pipeline}.

\subsection{3D People Detection}
\subsubsection{3D Detection}
The input stage requires the pointcloud data produced by the vehicle's lidar sensor to be processed, and features within the space identified. While this can take many forms, a simple clustering and concave hull algorithm is used to determine the raw locations of objects as a set of polygons. As an indiscriminate object detector, the pipeline groups all nearby points together into anonymous clusters with very low latency. However, this method makes no distinction between static and dynamic objects (such as poles and people). Hence, navigation schemes would need to make conservative decisions and assume all objects could potentially be pedestrians. This leads to overly conservative driving that is highly impractical in realistic scenarios, thus motivating the need for further downstream processing.

\subsubsection{2D Detection}
For 2D Detection, a people detector is used to operate on the RGB images from the onboard vehicle cameras. For this investigation, the detector used is YOLOv3 as described in Section \ref{ssec:yolo}. This design choice was motivated by fast runtime, acceptable performance and ease of implementation. 

In this module, the detector outputs a list of bboxes that outline where the system believes a person is present in the image. The coordinates for these bboxes are in UV-image coordinates, which is relative to the size of the image frame.

\subsection{Fusion}
\subsubsection{Pointcloud Projection}
To fuse the received pointcloud clusters with the received people bboxes, they must share a common reference frame. A function is used to project the pointclouds onto the images, the UV-image coordinate system was chosen to combine the two sensor modalities. This meant additional UV coordinate information was encoded into each point, resulting in a XYZUV pointcloud.

\subsubsection{Target Correlation and Verification}
With the XYZUV clusters, each point is overlayed with the `person' bboxes onto an image and compared for overlaps. If the overlap between one given cluster and another given bbox is sufficient, then that cluster is `validated' as a pedestrian in 3D. However, for each bbox, only one cluster can be validated, so only the best match is used. 

\begin{figure}[htbp]
    \centerline{\includegraphics[width=0.8\linewidth]{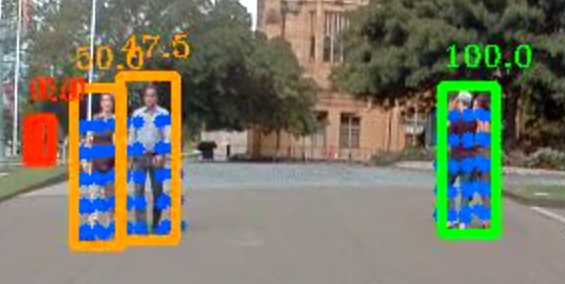}}
    \caption{Multiple People from 1 Cluster}
    \label{multi_detect}
\end{figure}

There are also situations where large groups of people may appear together, causing the clustering algorithm to bundle all these points together. The validation scheme handles this by assessing the best match in terms of the raw number of points overlapped and percentage of points in the cluster overlapped. Thus, one cluster may validate several bboxes, as can be seen in Figure \ref{multi_detect} on the left.

Algorithm \ref{alg:classify_polygons} outlines the above steps.

\begin{figure}[htbp]
    \centerline{\includegraphics[width=\linewidth]{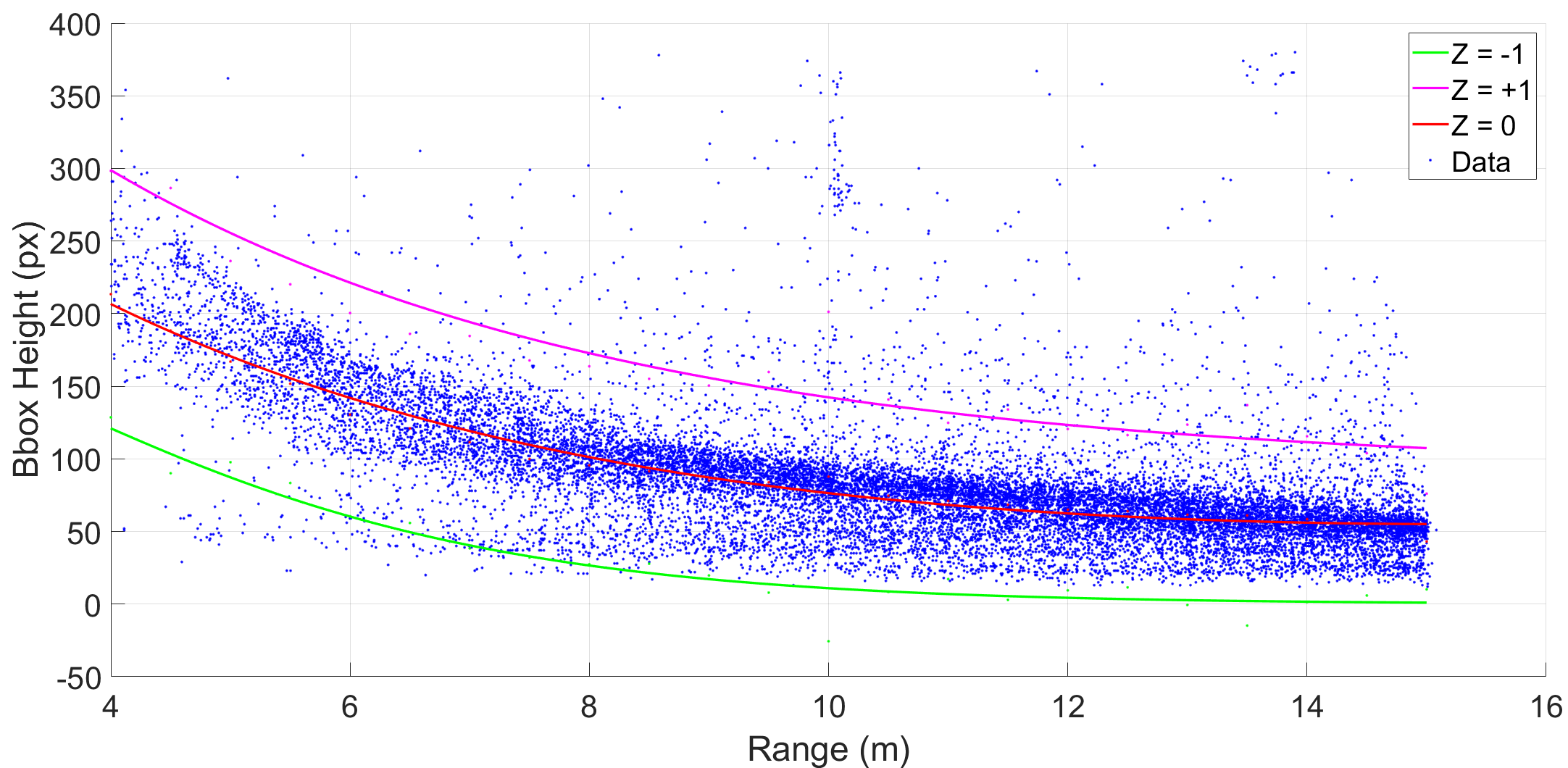}}
    \caption{Height v Range LOBFs for Sanity Check}
    \label{range_check}
\end{figure}

Finally, a sanity check is implemented on the system to filter out erroneous matches. For each match, a range to target and height of bbox was obtained. A statistical analysis of these values was performed to identify a trendline and reject outliers beyond 2 standard deviations of the mean. The results of this can be seen in Figure \ref{range_check}.

\begin{algorithm}
\DontPrintSemicolon 
\KwIn{\\$Bboxes$: A set of $n$ bboxes in UV coordinates\\ $Clusters$: A set of $k$ clusters of varying number of 3D points with UV projected coordinates}
\KwOut{\\Boolean $True$ or $False$}
$best\_overlap \gets 0$\;
$best\_cluster \gets None$\;
$overlapped\_points \gets None$\;
\For{$bbox$ in $Bboxes$}{
    \For{$cluster$ in $Clusters$}{
        $overlap, overlapped\_points \gets getOverlap(bbox, cluster)$\;
        \If{$overlap > best\_overlap$}{
            $range \gets cluster.range$\;
            \If{$bboxInBounds(bbox, range)$}{
                \uIf{$overlap > 0.8$}{
                    $best\_cluster \gets cluster$\;
                }
                \ElseIf{$overlap > 0.3$}{
                    $best\_cluster \gets overlapped\_points$
                }
                $best\_overlap \gets overlap$
            }
        }
    }
}
\caption{{\sc classifyPolygons}: matches bboxes with 3D clusters to detect pedestrians in 3D}
\label{alg:classify_polygons}
\end{algorithm}

\subsection{Calculating Proximity Layer Speed}
The 3D positions are used to identify the proximity of pedestrians to the intended vehicle path (approximated to project forward from the current vehicle wheel angle). While pedestrians on the path clearly need to slowed down for, how to handle those off the expected trajectory is more complicated. This is predominantly because pedestrian behaviour may be erratic, causing them to suddenly enter the path of the vehicle.

Thus, this work proposes a system where any detected people not in the direct path of the vehicle are projected forward using a tunable `lateral scaling factor'. This variable dictates the multiplying factor when converting lateral distance from the path to range down the path. For example: with a scaling factor of 3 and if there is a detected person 3 m from the vehicle path, the system will treat this individual as being 9 m down the path and then perform the calculation. For higher scaling factors, the system is less conservative and effectively only considers targets in the intended path. For lower scaling factors, the opposite is true. The Proximity layer speed can then be calculated for a given time-to-collision with these ranges along the path.

\section{Experiments and Results}
The previously outlined system was tested and validated against data from the USyd dataset. A variety of pedestrian densities was seen for each section of the route over multiple datasets. This allowed for more comprehensive performance evaluation.

\subsection{Context Layer}
\begin{figure}[ht]
    \centering
    \subfloat[][Shared/Semi-Shared Spaces\label{speed_profile_shared}]
        {\includegraphics[height=38mm]{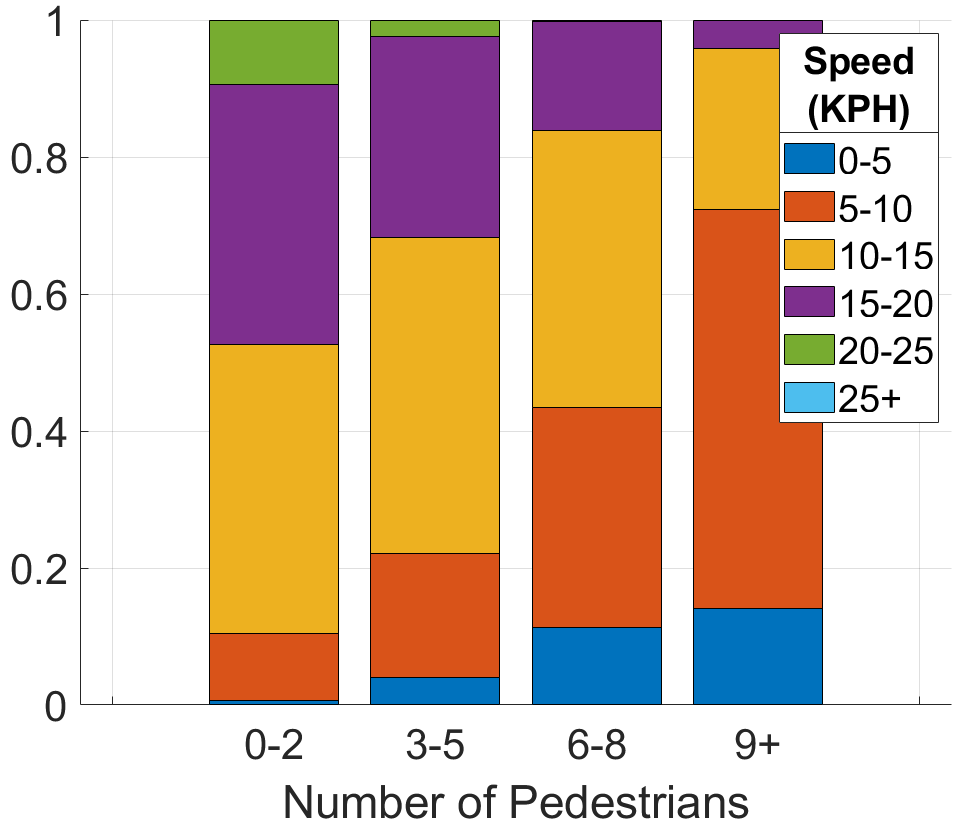}}
    \hspace{2mm}
    \subfloat[][Regular Roads\label{speed_profile_misc}]
        {\includegraphics[height=38mm]{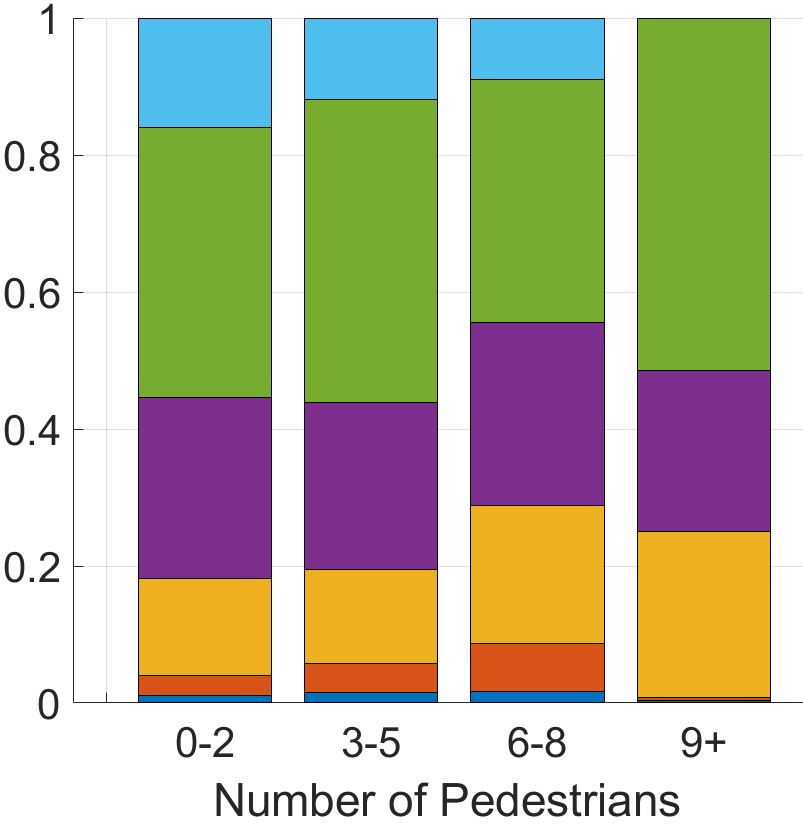}}
    \caption{Normalised Speed Profiles}
    \label{speed_profile}
\end{figure}

Figure \ref{speed_profile} shows the resultant stacked bar plots after tallying informative sections of 9 USyd datasets, resulting in over 30,000 data points. Bin widths of multiples of 3 for Pedestrian Density and steps of 5 KPH for Speed were chosen to capture the impact of small changes in pedestrian density on the overall speed of the vehicle.

The stacked bar graph is presented for visual evaluation of results, whereas the actual speeds used by the Speed Controller are taken as the average of each pedestrian density bin, shown in Table \ref{tab:speed_profile}\footnote{Theoretically, 0 pedestrian density would mean the first layer (legal speed limit) would be employed. Since this work did not consider this layer, it was omitted and grouped into the 0-2 bin.}.

\begin{table}[H]
    \centering
    \begin{tabular}{|c|c|c|}
        \hline
        \multicolumn{1}{|c|}{\multirow{2}{*}{Pedestrian Density}} & \multicolumn{2}{l|}{Context Speed (KPH)}\\
        \cline{2-3} \multicolumn{1}{| c |}{} & Shared & Regular \\
        \hline
        0-2 & 14.7 & 20.0\\
        3-5 & 13.0 & 19.7\\
        6-8 & 11.1 & 18.2\\
        9+  & 8.5 & 18.8\\
        \hline
    \end{tabular}%
    \vspace{6pt}
    \caption{Pedestrian Density to Speed Lookup}
    \label{tab:speed_profile}
\end{table}

From the above figures, the initial hypothesis was confirmed and correlation exists between vehicle speed and pedestrian density. The relationship found was that the appropriate speed is inversely related to the number of visible pedestrians. A much stronger correspondence can be seen in Shared and Semi-Shared Spaces than Regular Roads, suggesting that a singular driving scheme is insufficient and a consideration of contexts is crucial for pedestrian comfort. Additionally, the average speeds of vehicles was found to be significantly different between the two road contexts, further motivating the need for different driving schemes. 

\begin{figure}[H]
    \centerline{\includegraphics[width=0.9\linewidth]{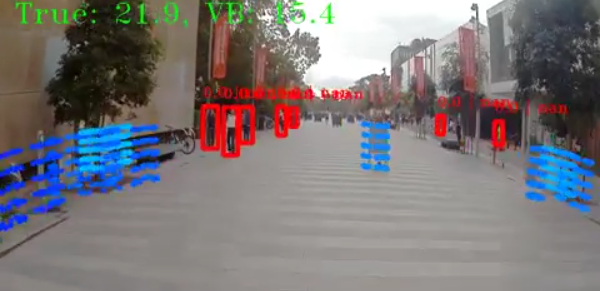}}
    \caption{Impact of Context Layer}
    \label{context_lim}
\end{figure}

Using Table \ref{tab:speed_profile}, the 2nd (Context) layer set appropriate speed limits for given contexts, as seen in Figures \ref{output2} and \ref{output1}, and also in Figure \ref{context_lim}. The latter image showcases how, despite no pedestrians being detected in 3D, the 2D detector still informs the system of the surrounding contexts and modulates down the speed. Overall, the Context layer was demonstrated to output vehicle speeds resembling a conservative human driver, suggesting that this approach to improving pedestrian comfort is effective. 

\subsection{Proximity Layer}
\begin{figure}[H]
    \centerline{\includegraphics[width=\linewidth]{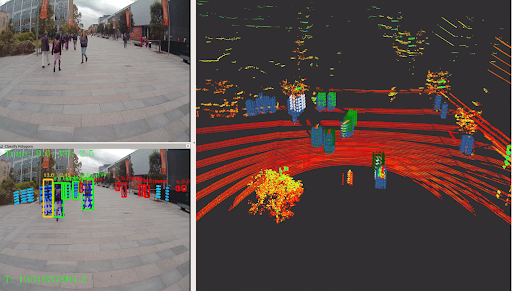}}
    \caption{Projected 3D Points overlapping with Bbox}
    \label{ped_detect}
\end{figure}

Figures \ref{multi_detect}, \ref{ped_detect} and \ref{context_lim} showcase the successful fusion of LIDAR and RGB detection methods into one common mode. Two types of outputs were used; one showing the percentage overlap (Figure \ref{multi_detect}) and another showing the range with z-score from pre-computed range expectations (Figures \ref{ped_detect}, \ref{context_lim}). Both of these metrics are used in conjunction for validating clusters, but were displayed separately for visual confirmation of expected system behaviour. 

Figure \ref{ped_detect} also shows the successful detection of people in 3D. The system can be seen behaving nominally as all people are highlighted in green markers whilst no other targets are highlighted. The rejection of misclassified clusters can also be seen functioning in the top left of the visualisation, highlighted by white markers. 

Importantly, this system also requires minimal additional runtime, taking under 10 ms to perform the fusion. Quality of performance was also adequate as pedestrians within the a range of around 15 m were reliably detected. It is noted however, that occasionally clusters in the background could be mis-matched with the bounding box of a person in the foreground, causing erroneous detections. This was deemed acceptable as these mis-detections only occurred in single frames between successful detections, hence no real target is practically missed (visible in the bottom, middle of Figure \ref{ped_detect}).

Insight into the 3D detector's behaviour was obtained by examining Figures \ref{output2} and \ref{output1}, as well as watching playbacks of the data. The disparity between `2D YOLO' and `3D People' detections is caused by the lower resolution of the LIDAR at longer ranges, leading to insufficient points for clustering. These missed detections are acceptable as higher ranged targets are generally accounted for by the Context layer.

Consequently, the Proximity layer was able to engage appropriately to restrict the speed only when pedestrians were in close proximity to the intended vehicle path. In situations where no pedestrians were detected, or none were in the path of the car, the output speed was undefined, displayed as gaps in the `Proximity' plots of Figure \ref{output2} and \ref{output1}. 

\subsection{Overall Performance}
\begin{figure*}[!ht]
    \centerline{\includegraphics[width=0.95\linewidth]{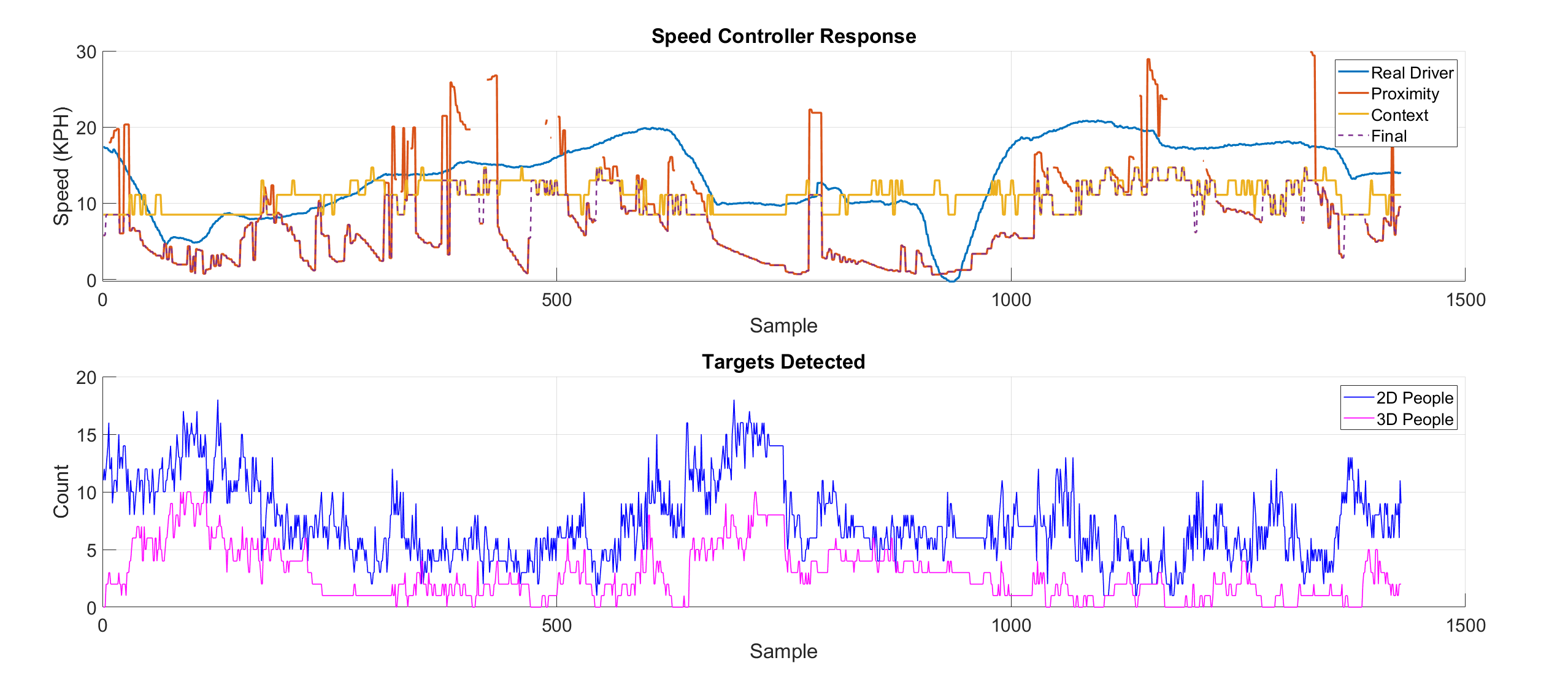}}
    \caption{Overall Output for Shared/Semi-Shared Spaces}
    \label{output2}
\end{figure*}

\begin{figure*}[!ht]
    \centerline{\includegraphics[width=0.95\linewidth]{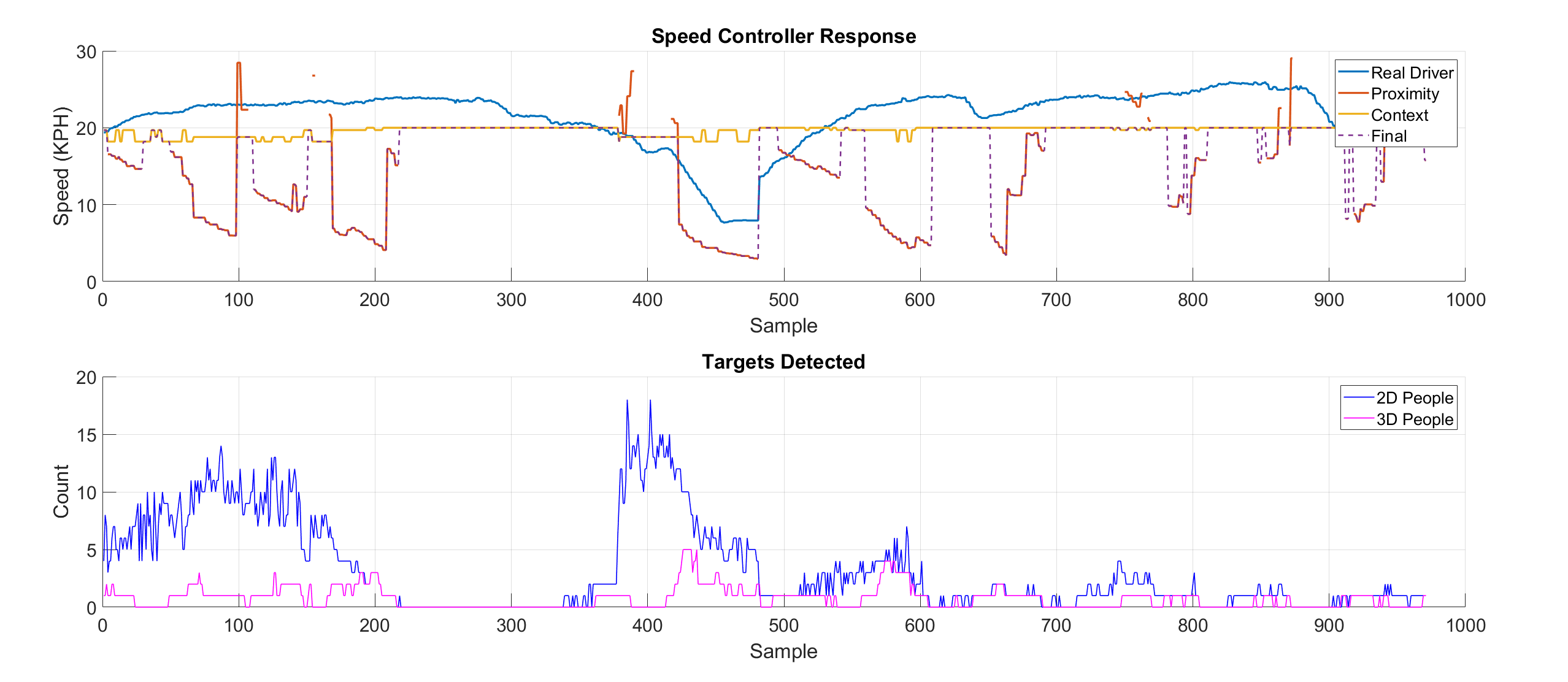}}
    \caption{Overall Output for Regular Roads}
    \label{output1}
\end{figure*}

The overall performance of the system was measured qualitatively by comparing the Speed Controller output with the vehicle speeds of a human driver. The overall results for one dataset are shown in Figures \ref{output2} and \ref{output1}. 

The top subplot of Figures \ref{output2} and \ref{output1} show the speed-time graphs and include the following values.  
\begin{itemize}
    \item Real Driver: Speeds from human driven vehicle.
    \item Proximity: Speeds from Fusion (3rd) Layer.
    \item Context: Speeds from Context (2nd) Layer.
    \item Final: Final speed output of the Speed Controller. Minimum value of all layers. 
\end{itemize}

In the lower subplot, the number of targets detected is also displayed to show how this value affects the behaviour of the system. 
\begin{itemize}
    \item 2D People: Number of people visible in the scene.
    \item 3D People: Number of people detected in the 3D fusion process.
\end{itemize}

In these figures, the need for multiple layers of control is highlighted as the Context and Proximity layers can be seen alternating to produce a `Final' speed that better resembles human driving. The overall behaviour of the system is found to be conservative, compared to the human driver, but maintains the general trend for both road contexts investigated. Therefore, the concept of layered speed control is successfully demonstrated. 

\begin{figure}[!htbp]
    \centering
    \subfloat[][Shared/Semi-Shared\label{hist2}]
        {\includegraphics[height=45mm]{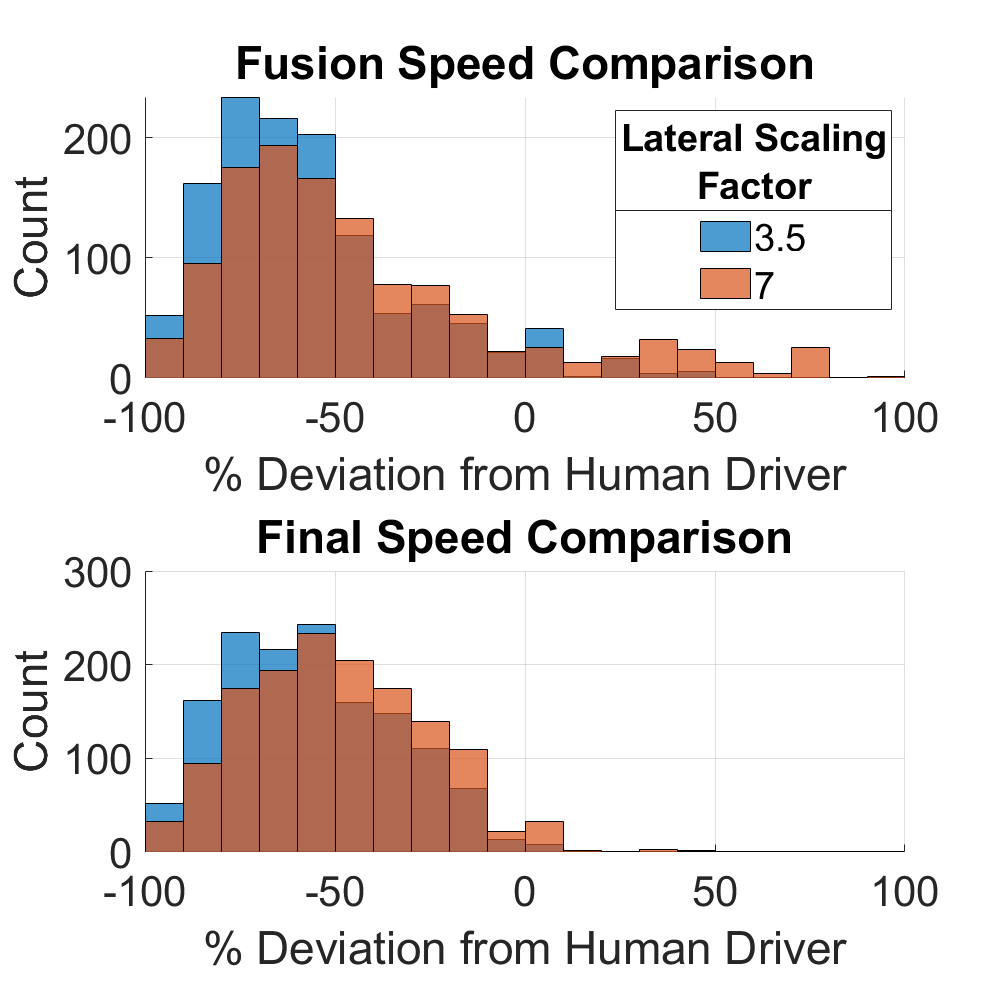}}
    \subfloat[][Regular Roads\label{hist1}]
        {\includegraphics[height=45mm]{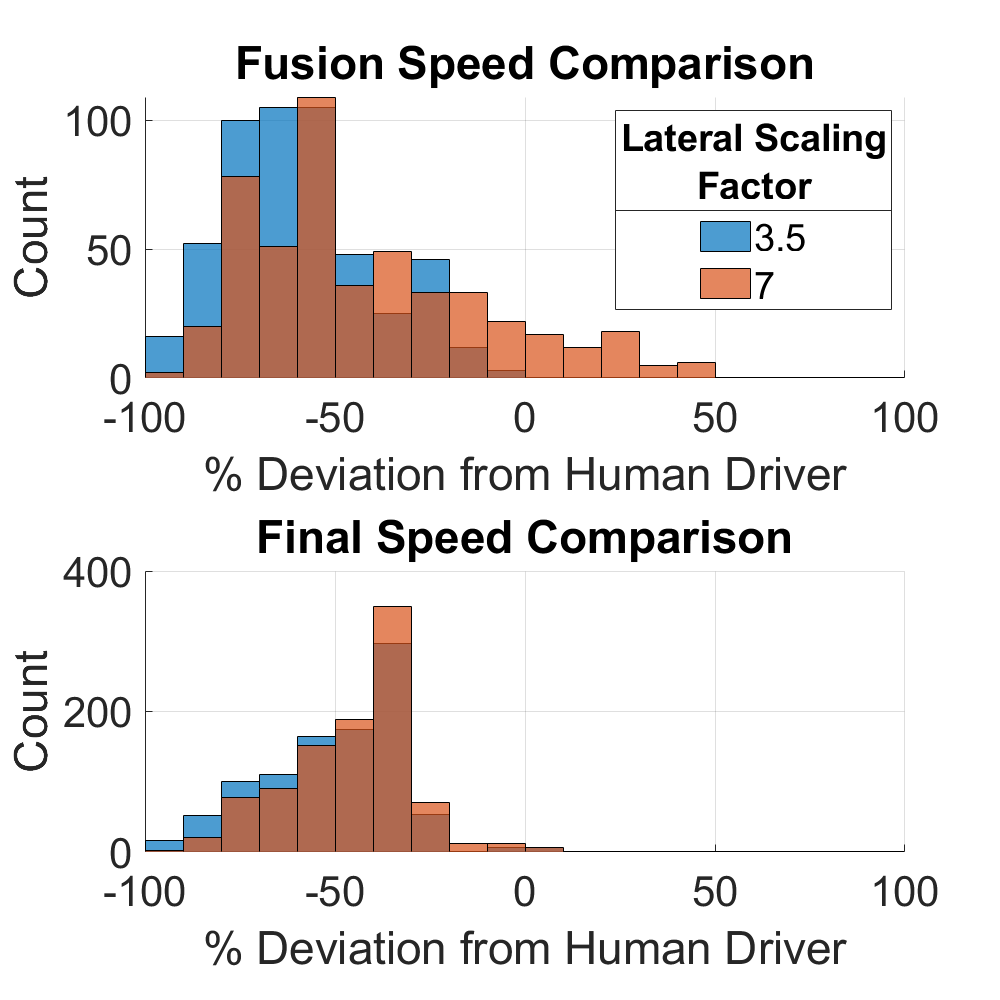}}
    \caption{Speed Distribution for 1 Dataset}
    \label{hist}
\end{figure}

\begin{figure}[H]
    \centerline{\includegraphics[width=0.85\linewidth]{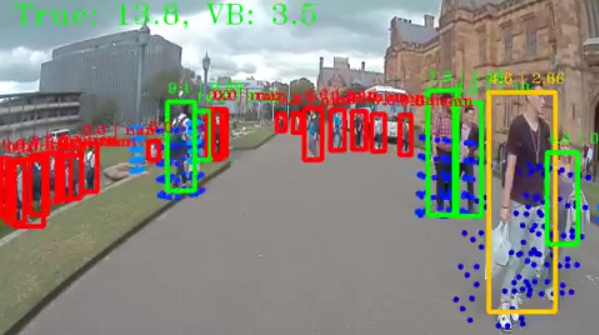}}
    \caption{Over-conservative System Behaviour}
    \label{overconservative}
\end{figure}

Figure \ref{hist} presents histograms of differences between human driven speeds and outputs from the implemented Speed Controller. These give a high level overview of how well the system mimics human behaviour, with values to the left of zero indicating more conservative performance, and vice versa. 
These histograms show that the system output matched human speeds more closely for regular roads. This was due to those sections being completely straight, such that the vehicle did not need to make slight turns to avoid obstacles. The inverse is true for shared and semi-shared spaces, as people were frequently directly in-front of the vehicle and forced evasive manoeuvres. The generally high pedestrian density lead to over-conservative Proximity layer speeds to be outputted. Figure \ref{overconservative} illustrates this behaviour, as despite there being no targets directly in front of the car, speed is heavily modulated. 

Figure \ref{hist} also shows how this conservative-ness can be moderated by adjusting the scaling factor. Predictably, higher scaling factors that projected targets further down the path caused the system to be less conservative and vice versa. This allows for such Contextual Speed Controllers to be tuned to better simulate human behaviour, although minimal tuning was done for the work in this paper. It is also noted that all of the showcased examples and datasets were around a university campus, hence there are inherently lower vehicle speeds. In comparing behaviour between low and high pedestrian density spaces, significant disparity was seen, suggesting that to navigate these spaces best, multiple driving schemes should be employed to adapt to the broader road context.

\section{Conclusions and Future Work}
The implemented Contextual Speed Controller successfully modulated the vehicle speed to mimic a conservative human driver. The results highlighted both the need for a layered approach when considering pedestrian comfort as well as the need for different driving schemes in different road contexts. The relationship between pedestrian density and vehicle speeds was also identified, quantified and utilised to modulate the speed in the experiments.

Overall, the implementation in this paper focused on ensuring pedestrian comfort and safety, as opposed to efficient navigation. Future improvements to the system can be made, particularly by improving the detection algorithms used. In particular, switching to a better performing 2D detector would improve the initial 2D bbox generation, which occasionally misses people due to inadequate processing times. Additionally, the system would benefit from knowing the planned vehicle trajectory, such that pedestrians not in the dynamic path of the vehicle can be omitted, reducing the conservative-ness of the system. Deploying the system and practically testing its performance on a vehicle whilst conducting user surveys would also be required to formally qualify the performance. 

\bibliographystyle{./bibliography/IEEEtran}
\bibliography{./bibliography/DJ.bib}

\end{document}